\documentclass[10.5pt,compsoc]{BD}
\usepackage{hyperref}
\hypersetup{hypertex=false,
            colorlinks=true,
            linkcolor=black,
            anchorcolor=black,
            citecolor=black}
\usepackage{graphicx}
\usepackage{footmisc}
\usepackage{subfigure}
\usepackage{url}
\usepackage{multirow}
\usepackage[noadjust]{cite}
\usepackage{amsmath,amsthm}
\usepackage{amssymb,amsfonts}
\usepackage{booktabs}
\usepackage{color}
\usepackage{ccaption}
\usepackage{booktabs}
\usepackage{float}
\usepackage{fancyhdr}
\usepackage{caption}
\usepackage{xcolor,stfloats}
\usepackage{comment}
\graphicspath{{figures/}}
\usepackage{cuted}  
\usepackage{captionhack}
\usepackage{epstopdf}
\usepackage{times}
\usepackage{afterpage} 
\usepackage{threeparttable}
\usepackage{graphicx}
\usepackage[numbers,sort&compress]{natbib}
\usepackage{picins}
\usepackage{lineno}
\usepackage{caption}  
\usepackage{diagbox}
\usepackage{booktabs}

\headevenname{\normalsize{\textbf{\emph{Big Data Mining and Analytics,} xxxxxxx}} 20xx, x(x): xxx-xxx}%
\headoddname{{\sf Junxiao Xue et al.:}\quad {\textbf{\emph{A Trustworthy Method for Multimodal Emotion Recognition ...}}}}%

\newtheoremstyle{mystyle}{0pt}{0pt}{\normalfont}{1em}{\bf}{}{1em}{}
\theoremstyle{mystyle}

\newcommand{\nop}[1]{}

\makeatletter
\renewcommand{\@biblabel}[1]{[#1]\hfill}
\makeatother

\makeatletter
\providecommand{\tst@headoddname}{}
\providecommand{\@thehead}{}
\makeatother

\usepackage{lineno}
\begin{document}


\thispagestyle{empty}

\clearpage

\hyphenpenalty=50000

\makeatletter
\newcommand\mysmall{\@setfontsize\mysmall{7}{9.5}}

\newenvironment{tablehere}
  {\def\@captype{table}}
  {}
\newenvironment{figurehere}
  {\def\@captype{figure}}
  {}

\thispagestyle{plain}%
\thispagestyle{empty}%

\let\temp\footnote
{}
\vspace*{-40pt}
\noindent{\normalsize\textbf{\scalebox{0.95}[1.0]{\makebox[5.9cm][s]
{BIG\hfill  DATA \hfill MINING \hfill AND \hfill ANALYTICS}}}}

\vskip .2mm
{\normalsize
\textbf{
\hspace{-5mm}
\scalebox{1}[1.0]{\makebox[5.6cm][s]{%
I\hfill S\hfill S\hfill N\hfill
~\hfill ~\hfill1\hfill0\hfill0\hfill7\hfill-\hfill0\hfill2\hfill1\hfill4
\hfill \quad {\color{white}{0\hfill x\hfill /\hfill x\hfill x\quad p\hfill p\hfill  x\hfill x\hfill x\hfill --\hfill x\hfill x\hfill x}\hfill}}}}

\vskip .2mm
{\normalsize
\textbf{
\hspace{-5mm}
\scalebox{1}[1.0]{\makebox[5.6cm][s]{%
DOI:~\hfill1\hfill0\hfill.\hfill2\hfill6\hfill5\hfill9\hfill9\hfill/\hfill B\hfill D\hfill M\hfill A\hfill.\hfill 2\hfill 0\hfill 2\hfill 5\hfill.\hfill 9\hfill 0\hfill 2\hfill 0\hfill 0\hfill 5\hfill 0}}}}

\vskip .2mm\noindent
{\normalsize\textbf{\scalebox{1}[1.0]{\makebox[5.6cm][s]{%
{V\hfill o\hfill l\hfill u\hfill m\hfill%
e\hspace{0.356em}x,\hspace{0.356em}N\hfill u\hfill%
m\hfill b\hfill e\hfill r\hspace{0.356em}x,\hspace{0.356em}%
x\hfill x\hfill x\hfill%
x\hfill x\hfill x\hfill x \hspace{0.356em}2\hfill0\hfill x\hfill x}}}}}\\

\begin{strip}
{\center
{\LARGE\textbf{
A Trustworthy Method for Multimodal Emotion Recognition}}
\vskip 9mm}

{\center {\sf \large
Junxiao Xue, Xiaozhen Liu\thanks{Corresponding author. Email: liuxiaozhen123@gs.zzu.edu.cn}, Jie Wang, Xuecheng Wu, and Bin Wu
}
\vskip 5mm}
%

\centering{
\begin{tabular}{p{160mm}}

{\normalsize
\linespread{1.6667} %
\noindent
\bf{Abstract:} { \sf
Existing emotion recognition methods mainly focus on enhancing performance by employing complex deep models, typically resulting in significantly higher model complexity. Although effective, it is also crucial to ensure the reliability of the final decision, especially for noisy, corrupted and out-of-distribution data. To this end, we propose a novel emotion recognition method called trusted emotion recognition (TER), which utilizes uncertainty estimation to calculate the confidence value of predictions. TER combines the results from multiple modalities based on their confidence values to output the trusted predictions. We also provide a new evaluation criterion to assess the reliability of predictions. Specifically, we incorporate trusted precision and trusted recall to determine the trusted threshold and formulate the trusted Acc. and trusted F1 score to evaluate the model's trusted performance. The proposed framework combines the confidence module that accordingly endows the model with reliability and robustness against possible noise or corruption. The extensive experimental results validate the effectiveness of our proposed model. The TER achieves state-of-the-art performance on the Music-video, achieving 82.40\% Acc. In terms of trusted performance, TER outperforms other methods on the IEMOCAP and Music-video, achieving trusted F1 scores of 0.7511 and 0.9035, respectively.}
}

\end{tabular}

}
\vskip 6mm

\vskip -3mm
\small\end{strip}

\thispagestyle{plain}%
\thispagestyle{empty}%
\makeatother
\pagestyle{tstheadings}

\begin{figure}[b]
\vskip -4mm
\small  
\renewcommand{\arraystretch}{0.9}  
\begin{tabular}{p{44mm}}
\toprule\\
\end{tabular}
\vskip -2.5mm
\noindent
\setlength{\tabcolsep}{1pt}
\begin{tabular}{p{1.5mm}p{79.5mm}}
$\bullet$& Junxiao Xue is with Research for Space Computing
System, Zhejiang Lab, Hangzhou, 311500, China.  E-mail: xuejx@zhejianglab.cn.\\
$\bullet$& Xiaozhen Liu is with the School of Cyber Science and Engineering, Zhengzhou University, Zhengzhou, 450001, China.  E-mail: liuxiaozhen123@gs.zzu.edu.cn.\\
$\bullet$& Jie Wang is with China Mobile (Hangzhou) Information Technology Co. Ltd., Hangzhou, 311100, China. E-mail: wangjie2@cmhi.chinamobile.com.\\
$\bullet$& Xuecheng Wu is with the School of Computer Science and Technology, Xi’an Jiaotong University, Xi’an, Shaanxi, 710049, China.  E-mail: wuxc3@stu.xjtu.edu.cn.\\
$\bullet$& Bin Wu is with the School of Computer and Artificial Intelligence, Zhengzhou University, Zhengzhou, 450001, China.  E-mail: wubin@gs.zzu.edu.cn.\\
$\sf{*}$& To whom correspondence should be addressed. \\
          &          Manuscript received: 2024-11-15;
          revised: 2025-2-15;
          accepted: 2025-04-28

\end{tabular}
\end{figure}



\vspace{3.5mm}

\section{Introduction}
\label{intro}

\noindent
Emotion recognition plays a crucial role in artificial intelligence, enabling machines to move beyond rigid scripts and respond more effectively to human emotions. Generally, emotion recognition methods can be categorized into two types. The first type is based on physiological signals, including electroencephalography (EEG)\textsuperscript{\cite{song2018eeg}}, electromyography (EMG)\textsuperscript{\cite{kulke2020comparison}}, and electrocardiography (ECG)\textsuperscript{\cite{jung2019utilizing}}. The second type is based on digital information, such as facial expression images\textsuperscript{\cite{tarnowski2017emotion, wang2022ease, li2020deep, wang2022d}}, body gestures\textsuperscript{\cite{noroozi2018survey}}, and speech signals\textsuperscript{\cite{latif2022self}}. As a data format that contains multiple types of digital information, videos provide a more comprehensive and accurate means of conveying emotion information, making them well-suited for research in emotion recognition. With the increasing availability of video datasets and the rapid development of video processing techniques, there has been a growing interest in analyzing emotion in videos.
 
In the field of emotion recognition, common multimodal methods typically model multimodal data by integrating multi-modality information with deep neural networks through feature fusion, facilitating emotion classification. However, this approach often overlooks the inherent uncertainties in emotion datasets. To this end, we introduce the concept of confidence to evaluate model performance more accurately from the trusted model\textsuperscript{\cite{han2022trusted}}. Specifically, we replaced the original information fusion module with the Confidence Module and the Belief Combination Module. In their research, Han et al.\textsuperscript{\cite{han2022trusted}} focused on mathematically evaluating the confidence of predictions and adopted a confidence-based fusion method for multimodal recognition results. 

Furthermore, while previous research\textsuperscript{\cite{han2022trusted}} has explored confidence methods, there remains a lack of credible evaluation metrics, limiting the research and application of confidence methods. Building on this, we combined the concept of trustworthiness with binary classification criteria\textsuperscript{\cite{davis2006relationship}} to propose a trusted method for loss calculation and several trusted evaluation metrics. 

In real-life scenarios, models may generate incorrect predictions due to the high ambiguity of the data. Therefore, it is essential to compute prediction results, perform loss calculation, and evaluate the effectiveness of classification results under high-confidence conditions. To this end, we
compared the model training process using the proposed
loss method with traditional loss methods on two datasets and evaluated the model performance, including
classification and trusted performance as evaluation
metrics. 
The results demonstrated the effectiveness of the proposed loss method in trusted model training and the rationality of the proposed evaluation method. To summarize, our significant contributions are threefold:

\begin{itemize}
    \item Constructed a Trusted Emotion Recognition(TER) model, a confidence  assessment method is also introduced, and the fusion of multimodal recognition results is realized based on this module. The model has achieved classification performance close to the state-of-the-art methods on two datasets.

    \item Optimized the training method of the model by incorporating confidence  assessment into the loss calculation. This can ensure that the model focuses more on the confidence of the results during the training process while maintaining classification performance and improving trusted performance.

    \item Proposed a trusted evaluation criterion, which incorporates confidence  into assessing the model's predictions. This criterion also provides guidelines for selecting a trusted threshold, supporting the application of trusted results. The proposed emotion recognition model has achieved state-of-the-art trusted performance on two datasets.
\end{itemize}

The remaining sections of this paper are organized as follows. Section \ref{relate} contains a brief survey of related works. The details of the proposed framework, loss calculation method and trusted evaluation criterion are given in Section \ref{method}. Section \ref{experiments} conducts the ablation experiments of the multimodal model and the loss method, compares the performance with existing models and verifies the generalizability of our proposed model. Finally, Section \ref{conclusion} presents concluding comments and future work.

\section{Related Work}\label{relate}
\noindent
First of all, we discuss the works on utilizing images and audio for emotion recognition and highlight some common challenges. Subsequently, we present studies in deep learning on confidence estimation, including the methods which can be adapted.

\subsection{Emotion Recognition}

It is a common approach to conduct emotion recognition by extracting image features and audio features. Classical studies\textsuperscript{\cite{li2018facial, asghar2020eeg}} conducted emotion recognition based on ResNet and VGG models.

\subsubsection{Emotion Recognition Challenges}

Based on the existing model structures, we can observe that the primary challenge lies in the temporal dimension. ResNet and VGG are 2D convolutional neural networks that primarily focus on spatial information. However, the most significant difference between video and image is the temporal information. Although the RNNs could extract temporal information, their convolutional kernels cannot extract spatial information. Therefore, the current challenge is constructing a deep model that simultaneously extracts spatio-temporal information from video and audio.

On the other hand, when using multimodal information, the appropriate fusion method plays a crucial role in enhancing model performance. Compared to early fusion methods, late fusion methods\textsuperscript{\cite{carreira2017quo, noroozi2017audio, priyasad2020attention}} can better leverage the excellent feature extraction capabilities of single-modal methods. In these studies, researchers deploy different methods to provide fixed fusion weights to the models. However, the information weights of different modalities in various datasets are disparate, and fixed weight methods can only fit the weights of the overall data while neglecting the outliers. Therefore, adaptive fusion methods are considered the optimal solution for late fusion, and current research mainly focuses on addressing this challenge.

\subsubsection{Emotion Recognition Methods}

With the rapid growth and success of deep learning applications in the past decade, there has been a substantial increase in the number of methods employed to conduct emotion recognition. From emotion recognition in natural language processing (NLP), RNNs are introduced to conduct emotion recognition. Subsequently, CNN-based methods are adopted for affective image content analysis(AICA)\textsuperscript{\cite{khaireddin2105facial}}, which is a commonly used approach in facial emotion recognition. 
In affective video content analysis (AVCA), Xue et al.\textsuperscript{\cite{xue2024affective}} have introduced advanced methods developed over the past decade for addressing video feature extraction, expressing subjectivity, and multimodal feature fusion.

The Transformer has gained attention from various research fields due to its outstanding performance and has been combined with methods in various fields to address challenges. The Swin Transformer\textsuperscript{\cite{liu2021swin}} enhanced the feature extraction capability of models by introducing attention mechanisms into two-dimensional convolutional networks, achieving state-of-the-art results in multiple image classification tasks. Wang et al.\textsuperscript{\cite{xiaohua2019two}} applied this method to solve AICA and outperform the original models. The improvement in performance by incorporating attention mechanisms into three-dimensional convolutional networks has also attracted the interest of more researchers. Lee et al.\textsuperscript{\cite{lee2019context}} proposed an AVCA architecture that combines facial emotion recognition and attention mechanisms. Recent developments in multimodal emotion recognition include the Frame-SCN framework\textsuperscript{\cite{shi2025multimodal}}, which utilized framelet transformations to decrease text dependency and boost accuracy. The GM2RC model\textsuperscript{\cite{shi2024gm2rc}} enhanced analytical precision through intra-modal refinement and intermodal complementation. Meanwhile, FrameERC\textsuperscript{\cite{li2025frameerc}} employd graph framelet transformations and a dual-reminder fusion mechanism, effectively capturing emotional nuances and amplifying the impact of non-textual modalities, leading to superior performance in ERC tasks.

In multimodal research, researchers have also recognized the impact of attention mechanisms. Applying attention mechanisms to the fused information enhances the capability of feature extraction. Gupta et al.\textsuperscript{\cite{gupta2018attention}} demonstrated the powerful performance of this approach in their proposed model. On the other hand, Priyasad et al.\textsuperscript{\cite{priyasad2020attention}} applied attention mechanisms to the late fusion process, adjusting the fusion weights during training to achieve optimal performance. Han et al.\textsuperscript{\cite{han2022trusted}} also designed a confidence module based on numerical features of prediction results and proposed an adaptive fusion method that uses confidence as fusion weights. Compared to fixed weight methods, this specific approach, which assigns different fusion weights for different groups, has shown superior performance in multiple performance metrics.

\subsection{Trusted Result}

Deep learning has been widely recognized and has achieved remarkable performance in research fields such as computer vision, natural language processing, and data mining. It has also been successful in practical applications such as object detection, speech recognition, medical diagnostics, and financial fraud detection. However, the inherent lack of interpretability and theoretical support in deep learning methods provided insufficient explanations for the results they produce. Consequently, the application of deep learning has been greatly limited, especially in risk-sensitive scenarios where it has yet to gain widespread practical use. Analyzing the mathematical characteristics of the likelihood-based results generated by deep learning methods has become a significant direction in studying the confidence of deep learning outcomes.

Jun Deng et al. proposed to use confidence to evaluate the recognition results of machine learning methods\textsuperscript{\cite{deng2012confidence1}}, and proposed a CM for a SER system based on semi-supervised learning\textsuperscript{\cite{deng2012confidence2}}. Research on confidence in deep learning initially stemmed from the work of Guo et al.\textsuperscript{\cite{guo2017calibration}}. They proposed to utilize confidence calibration to constrain the confidence of model predictions within an optimal range. The confidence calibration was analyzed for various state-of-the-art models in the computer vision and natural language processing domains across different datasets. Experimental results indicated that the confidence of most current deep learning models tends to be overconfident, with predicted confidence levels exceeding their accuracy. Since then, extensive research has been conducted on the confidence of deep learning models. Mukhoti et al.\textsuperscript{\cite{mukhoti2020calibrating}} attributed the poor confidence calibration to over-parameterization in deep neural networks, enabling models to memorize the entire training set and maximize confidence for all samples. However, research by Bai et al.\textsuperscript{\cite{bai2021don}} demonstrated that even the simplest logistic regression models still exhibited overconfidence in datasets, revealing that the confidence calibration ability of models is independent of the number of model parameters. Wang et al.\textsuperscript{\cite{wang2021confident}} focused on practical applications and introduced various methods for confidence calibration in deep models and graph neural networks. Yang Li et al.\textsuperscript{\cite{li2022confidence}} applied the confidence approach on the speech emotion recognition task and proposed EmoConfidNet to obtain excellent model performance.

\begin{figure*}[t]
    \centering
    \includegraphics[width=\linewidth]{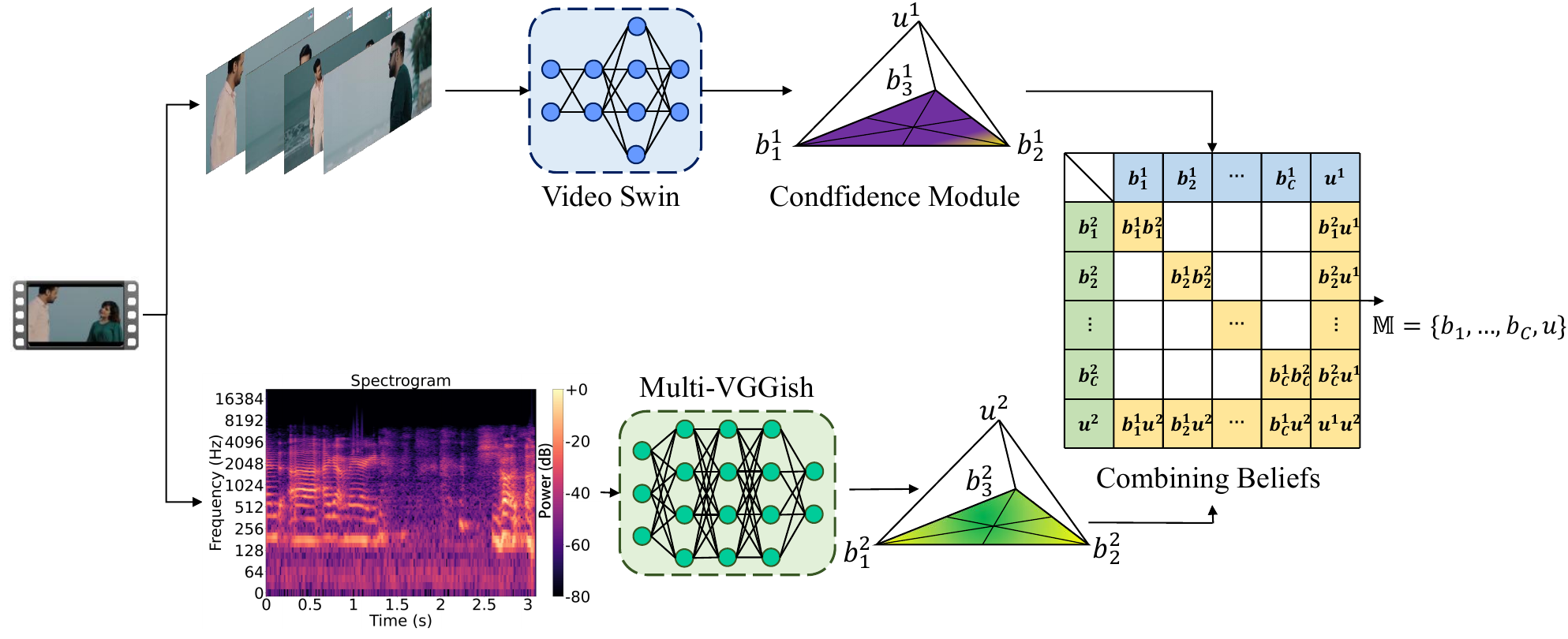}
     \caption{\textbf{The overall illustration of our proposed TER model. The model consists of four main modules, which are Video Swin-Transformer Module, Multi-VGGish Module, Confidence Module, and Combining Beliefs Module, respectively.}}
    \label{TER}
\end{figure*}

Research on confidence in multimodal models primarily focused on the confidence of the final results, employing similar methods for confidence calibration as those used in single-modal models. Han et al.\textsuperscript{\cite{han2022trusted}} proposed a late fusion method based on confidence in their research. The distinct feature of this method was that Han et al. assessed the confidence of different single-modal results, utilizing these confidences as fusion weights to achieve confidence fusion. They redesigned the confidence module by modeling the confidence of the model results using the Dirichlet distribution. This approach provided confidence that reflected the uncertainty of the model's outputs. Dempster-Shafer evidence theory was introduced for confidence fusion, resulting in fused outcomes that incorporated uncertainty. This fusion method possessed solid theoretical support and a set of adaptive fusion methods based on confidence. Extensive experiments have demonstrated the advantages of this fusion method, validating its accuracy, robustness, and effectiveness.

\section{Method}\label{method}

\noindent
In this section, we address a key limitation of conventional multimodal emotion recognition methods, which often neglect the uncertainty inherent in multimodal emotion datasets. To overcome this, we extend the existing trusted model\textsuperscript{\cite{han2022trusted}} to the field of emotion recognition and introduce the Trusted Emotion Recognition (TER) model for affective video content analysis (AVCA). Furthermore, previous trusted models\textsuperscript{\cite{han2022trusted}} did not consider the trustworthiness of loss functions and evaluation metrics. As a result, we present a comprehensive explanation of the loss function method and trusted evaluation criteria developed specifically for this model.

\subsection{Trusted Emotion Recognition Model}

In previous work, decision-making typically utilized traditional multimodal information fusion approaches. Although multimodal models using information fusion can provide accurate classification outcomes, they often result in unreliable predictions, especially when there is a conflict between the recognition results of the two modalities. Our method builds upon decision level fusion (late fusion) techniques. Traditional decision level fusion methods integrate different classification results using fixed weights, often overlooking the uncertainties inherent in decision-making. In contrast, our approach considers the confidence of each classification result, dynamically adjusting the weights of these results. Compared to traditional decision level fusion methods, our method achieves higher Trusted accuracy, indicating improved prediction precision under high confidence levels. Therefore, we propose a new confidence-based multimodal architecture for performing affective video content analysis.

We propose the Trusted Emotion Recognition (TER) model, which consists of four main modules, as shown in Fig. \ref{TER}. The first module is the Video Swin-Transformer Module\textsuperscript{\cite{liu2022video}}, which extracts emotion-related features from the video frame sequence and performs emotion classification for the video modality. The second module is the Multi-VGGish Module\textsuperscript{\cite{hershey2017cnn}}, which extracts emotion-related features from the audio and performs emotion classification for the audio modality. The third module is the  Confidence Module, which measures the confidence of each modality's classification results. The fourth module is the Combining Beliefs Module, which fuses the classification results of each modality based on their confidence to obtain the final multimodal emotion recognition predictions.

\begin{figure*}[t]
    \centering
    \includegraphics[width=\linewidth]{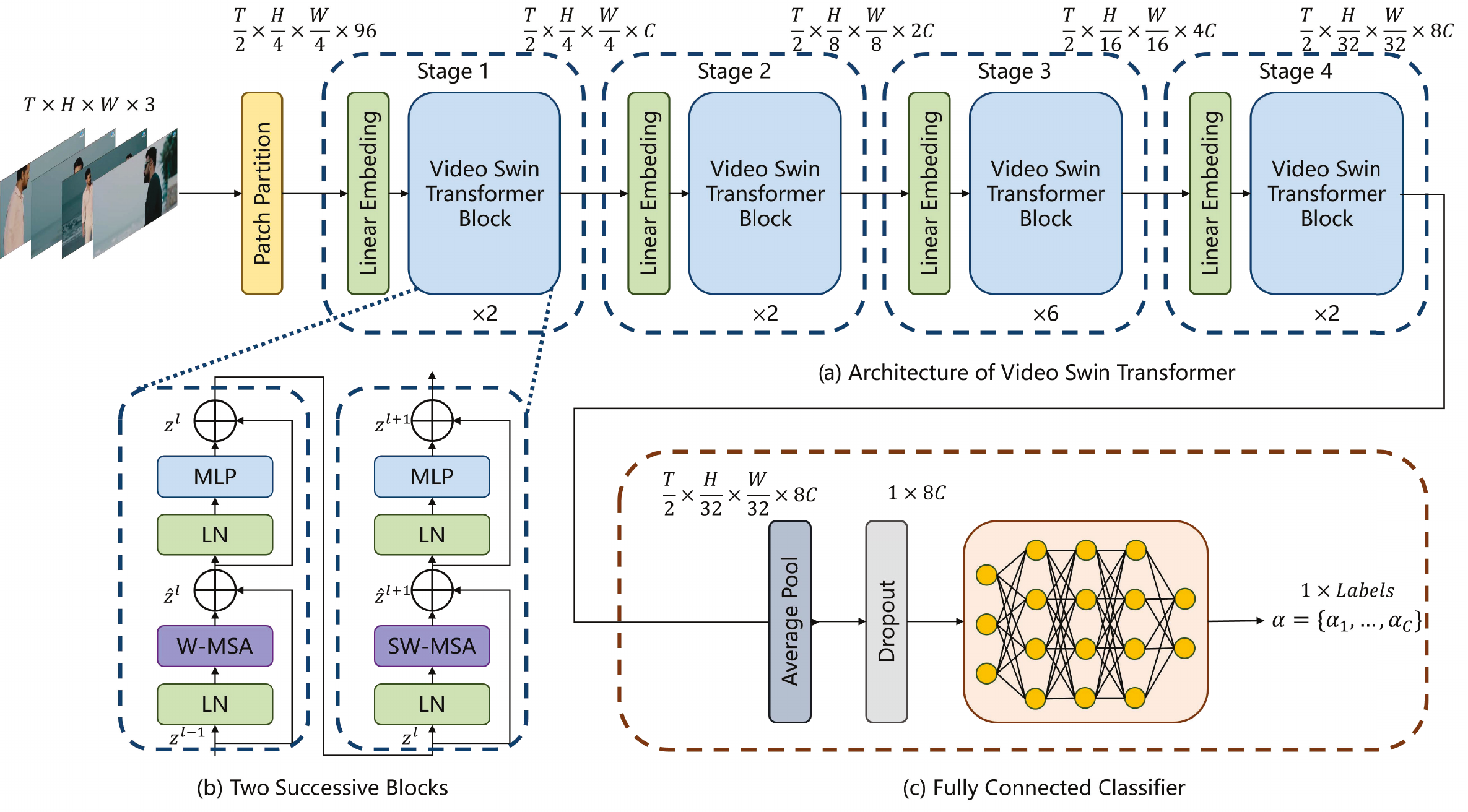}
    \caption{\textbf{The overall illustration of the visual stream in our proposed TER model. (a) shows the architecture of Video Swin Transformer \cite{liu2022video}. Specifically, the Two Successive Blocks and the Fully Connected Classification are shown as (b) and (c), respectively.}}
    \label{vswin}
\end{figure*}

\begin{figure*}[t]
    \centering
    \includegraphics[width=\linewidth]{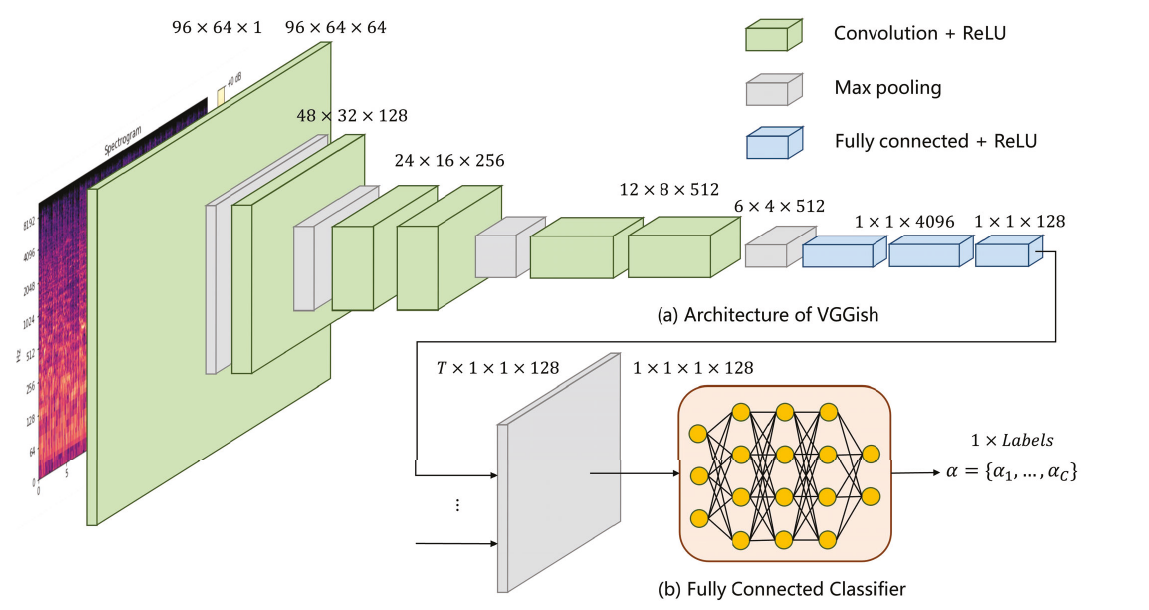}
    \caption{\textbf{The overall illustration of the audio stream in our proposed TER model. Specifically, the prediction of VGGish\textsuperscript{\cite{hershey2017cnn}} is input to the Fully Connected Classification to output the final predictions.}}
    \label{vggish}
\end{figure*}

\subsubsection{Video Swin-Transformer Module}

Visual information is the primary way humans receive information, encompassing rich emotional expressions such as facial expressions and body movements. However, traditional 2D convolutional methods, such as MAE\textsuperscript{\cite{he2022masked}} and MAGE\textsuperscript{\cite{li2023mage}}, tend to lose important temporal features when processing images. To this end, we select frame sequences containing spatial and temporal information in the proposed multimodal architecture as the source of image modality features. We introduce the Video Swin-Transformer\textsuperscript{\cite{liu2022video}} with the specific attention mechanism to extract spatio-temporal features from frame sequences.

The structure of the Video Swin-Transformer is illustrated in Fig.\ref{vswin}. First, the video frame sequence undergoes Patch Partition and Linear Embedding, which divide input into a series of spatio-temporal blocks and embed them. The feature extraction module reduces dimensionality and extracts features through four stages. The Swin Transformer Block employs an attention mechanism within each stage to extract spatio-temporal features from the input. Compared to traditional Transformer models, this block restricts the attention calculation within a window and introduces a 3D relative positional bias matrix $B \in R^{P^{2} \times M^{2} \times M^{2}}$ for each head. The attention calculation equation is as follows:
\begin{equation}
    \normalsize
    \mathrm{Attention}(Q, K, V) = \mathrm{Softmax}\left(\frac{QK^{\mathrm{T}}}{\sqrt{d}} + \mathrm{B}\right)V,
\end{equation}
where $Q, K, V \in R^{PM^{2} \times d}$ are the \textit{query}, \textit{key}, and \textit{value} matrices, respectively; $d$ is the dimension of \textit{query} and \textit{key} features. $P, M, M$ represent the size of the 3D window, and $PM^{2}$ is the number of tokens in a 3D window. Two consecutive Block structures are employed in each stage, as depicted in Fig.\ref{vswin} (b). Except for the last stage, Patch Merging is applied at the end of each stage to perform 3D dimension reduction on the extracted features. Finally, a fully connected classifier maps the features to the classification results, as shown in Fig.\ref{vswin} (c).

\subsubsection{Multi-VGGish Module}

Audio is the primary way humans convey information, using language to express ideas and emotions, such as relaxed humming, joyful cheers, angry shouts, fearful screams, and sorrowful sighs. The mature audio feature extraction method VGGish\textsuperscript{\cite{hershey2017cnn}} has achieved remarkable performance on various kinds of datasets. We employ this model as the baseline for the audio module and construct our Multi-VGGish module that can extract the spatio-temporal features of audio, enabling the classification of multiple seconds of input data.

The structure of Multi-VGGish is shown in Fig.\ref{vggish}. In this module, the input is a sequence of mel-spectrograms, which are sliced along the time dimension and fed into the VGGish sub-modules. After feature extraction in each VGGish sub-module, the extracted features are inputted into the feature fusion sub-module and combined along the time dimension. To robustly handle the input, we analyze the data format in the dataset and pad the data that does not meet the time dimension requirements. Therefore, before classification, we add a max pooling layer along the time dimension to reduce dimensions and eliminate the impact of padding data. 

Subsequently, we employ a fully connected layer to classify the features, and the structure of the classifier is shown in Fig.\ref{vggish} (b).

\subsubsection{Confidence Module}

We need a unimodal confidence method to achieve confidence-based multimodal fusion. Although the output of softmax is often considered as the prediction confidence, existing research has shown that such confidence can lead to overconfidence issues. To obtain unimodal confidence value, Han et al.\textsuperscript{\cite{han2022trusted}} replaced the softmax layer of the model  with a softplus layer and obtained the Dirichlet distribution for the model's classification outputs $\alpha =\{\alpha_1, \ldots ,\alpha_C\}$ using the following method, C denotes the total number of classes, and 
e represents the evidence.:
\begin{equation}
    e = \mathrm{Softplus}(\alpha) + 1,
\end{equation}
based on the proposed confidence theory, the belief mass $b$ and confidence value $u$ are calculated as follows:
\begin{equation}
    b_c = \frac{e_c - 1}{\sum_{c=1}^{C} e_c} \quad \mathrm{and} \quad u = \frac{C}{\sum_{c=1}^{C} e_c}.
\end{equation}

The confidence value $u$ can also be understood as the uncertainty mass. \( u_1 \) represents the uncertainty of the visual information, and  \(u_2\) represents the uncertainty of the audio information. The lower the uncertainty, the higher the confidence of the prediction. We implement our confidence module based on this theory. Different from the multimodel fusion method, we combine the belief mass \( b \) with the confidence value \( u \) as the module's trusted results, as shown in the Confidence module of Fig. \ref{TER}. In Fig. \ref{TER}, for \( b^{m}_{n} \), the superscript \( m \) represents different modalities, where 1 refers to visual information and 2 refers to audio information. The subscript \( n \) indicates the various categories in emotion classification. Through the confidence module, the uncertainty \( u \) of the results from different modalities is finally obtained.

In model performance evaluation, confidence is the core of our model, directly participating in calculating various metrics and demonstrating the effectiveness of a trusted multimodal model.

\subsubsection{Combining Beliefs Module}

According to the uncertainty and evidence theory proposed by Han et al.\textsuperscript{\cite{han2022trusted}}, we can obtain unimodal results composed of classification evidence and uncertainty. When fusing these results, we need to use the Evidence Theory. Based on the DS-Combination Rule proposed by Han et al.\textsuperscript{\cite{han2022trusted}}, the fusion of classification results follows the below rules:
\begin{equation}
    \normalsize
    b_c = \frac{1}{1-k} (b_c^1 b_c^2 + b_c^1 u^2 + b_c^2 u^1) \quad \mathrm{and} \quad u = \frac{1}{1-k} u^1 u^2,
\end{equation}
where $k = \sum_{i \neq j} b_i^1 b_j^2$ is a measure of the amount of conflict between the two mass sets, and the scale factor $\frac{1}{1-k}$ is used for normalization. 
Conventional decision-level fusion methods typically use fixed weights to select the analysis results from visual and audio modalities. In contrast, our method integrates uncertainty \( u \) into the fusion process using a simplified Dempster-Shafer combination rule. Refer to the Combining Beliefs module in Fig. \ref{TER}. This allows for dynamic adjustment of weights across different modalities, better handling inconsistencies between the recognition results of the two modalities, and improving recognition accuracy.
The belief mass $b$ and confidence value $u$ are combined into the trusted result.

\subsection{Trusted Cross-Entropy Loss}

We work on obtaining trusted multimodal classification results by joint training neural networks. In traditional neural network classifiers, cross-entropy loss is widely deployed. 

In our model, the output predictions include the belief mass  \( b_i \), which can be understood as the probability of each class, and \( u_i \), which represents the confidence value for the other class, indicating the inability to confidently identify the result.
The model's prediction for the i-th sample is represented as $\hat{Y}_i = \{\hat{b}_{i1}, ..., \hat{b}_{iC}, \hat{u}_i\}$, and it satisfies $\sum_{c=1}^C \hat{b}_{ic} + \hat{u}_i = 1$. The ground-truth emotion labeling is denoted as $Y_i = \{b_{i1}, ..., b_{iC}, u_i\}$, and it follows $\sum_{c=1}^C b_{ic} + u_i = 1$. The trusted cross-entropy loss function can be expressed as:
\begin{equation}
    \label{eq_loss}
    \mathcal{L}_{t\_ce}(\hat{Y}_i; Y_i) = -\sum_{c=1}^{C} b_{ic} \log \hat{b}_{ic} - u_i \log \hat{u}_i.
\end{equation}

Considering the influence of different modalities during multimodal training, we adopt a multi-task strategy to collect results from all modalities to improve fusion performance. We employ three loss functions to jointly train the TER model in an end-to-end manner:
\begin{equation}
    \mathcal{L}^{\mathrm{overall}} = \mathcal{L}_{t\_ce}^v + \mathcal{L}_{t\_ce}^a + \mathcal{L}_{t\_ce}^c,
\end{equation}
where $\mathcal{L}_{t\_ce}^v$ represents the loss for the video branch, $\mathcal{L}_{t\_ce}^a$ represents the loss for the audio branch, and $\mathcal{L}_{t\_ce}^c$ represents the loss for the fusion results.

\subsection{Trusted Evaluation Criterion}
\label{t_evaluation}

Common evaluation metrics such as accuracy, precision, and F1 score do not take into account the concept of confidence, which can lead to unreliable predictions when modality data is ambiguous. Similarly, when humans classify objects, they also assess the credibility of the classification results. By employing a trusted model that utilizes Trusted Accuracy, it is possible to compute the predictive accuracy under conditions of high confidence, making this approach practically applicable. In the inference process, our proposed evaluation metrics include not only the evaluation metric results but also the uncertainty associated with these results. This uncertainty is compared to a confidence threshold; if the calculated uncertainty is below this threshold, the result is deemed reliable; otherwise, it is considered unreliable. Therefore, we have defined trusted evaluation criteria suitable for trusted models in this subsection. These evaluation criteria will be validated through experiments, demonstrating the superiority of trusted models compared to regular models.

\subsubsection{Trusted Classification}

For classification tasks, the evaluation is typically limited to the correctness of the classification results. However, we consider the classification results and the confidence in the trusted results. When evaluating a trusted model, it is essential to assess these two aspects comprehensively to obtain a more comprehensive evaluation. The trusted results can be divided into true or false classifications, while the confidence of trusted results can be categorized as high or low. Then we consider both aspects and utilize a confusion matrix in Table \ref{confusionm} to represent all trusted results. In this table, $\mathrm{HT}$ denotes a trusted result that is correctly classified with high confidence, $\mathrm{LT}$ denotes a trusted result that is correctly classified with low confidence, $\mathrm{HF}$ denotes a trusted result that is incorrectly classified with high confidence, and $\mathrm{LF}$ denotes a trusted result that is incorrectly classified with low confidence. \(N\) denotes the total number of samples. In the subsequent evaluation criteria, we will deploy this evaluation result for representation.
\begin{table}[h]
\centering
\begin{tabular}{c ccc}
\toprule
\multirow{2}{*}{\shortstack{Confidence}} & \multicolumn{3}{c}{Classified} \\
\cmidrule(lr){2-4}
 & TRUE & FALSE & Total \\
\midrule
High & HT & HF & HT+HF \\
Low  & LT & LF & LT+LF \\
\midrule
Total & HT+LT & HF+LF & $N$ \\
\bottomrule
\end{tabular}
\caption{\textbf{The confusion matrix we proposed in the trusted classification.}}
\label{confusionm}
\end{table}



\subsubsection{Trusted Threshold}

There are multiple methods to determine the correctness of classification results. When the number of classes is small, the highest probability label consistent with the true label can be used as the criterion. However, when the number of classes is large, it is common to use whether the true label is included in the top-$K$ classification probability results as the criterion, where $K$ is a small number ($e.g.$, TOP 5). In our experiment, the classes in the dataset are few (less than ten), so we will use the first criterion, whether the highest probability label is consistent with the true label, as the judgment basis.

Determining the confidence threshold (the degree of uncertainty) requires more consideration and analysis. The trusted results generated by the model only include a confidence value, and the threshold cannot be determined through internal comparison. The study of binary classification results\textsuperscript{\cite{davis2006relationship}} provides some insights: different classification thresholds for the same model will result in different classification performances. Setting a reasonable classification threshold is necessary to obtain the best binary classification performance. In the trusted results, changes in the confidence threshold will also affect model performance. Therefore, we introduce the concepts of precision and recall to evaluate the performance of trusted results and determine a reasonable confidence threshold by balancing these two indicators, thereby achieving excellent model performance.   As the trusted threshold increases, the confidence value \( u \) increases, while the trusted precision increases and the trusted recall decreases.
 Trusted precision $(\mathrm{TP})$ represents the ratio of correctly classified results in high-confidence predictions, which is defined as:
\begin{equation}
\label{eq_tp}
    \mathrm{TP} = \frac{\mathrm{HT}}{\mathrm{HT} + \mathrm{HF}},
\end{equation}
where trusted recall $(\mathrm{TR})$ represents the ratio of high-confidence results in the correct predictions, which is defined as:
\begin{equation}
\label{eq_tr}
    \mathrm{TR} = \frac{\mathrm{HT}}{\mathrm{HT} + \mathrm{LT}},
\end{equation}
We can comprehensively assess the model's performance on trusted results by calculating trusted precision and trusted recall.

\begin{figure}[h]
    \centering
    \includegraphics[width=\linewidth]{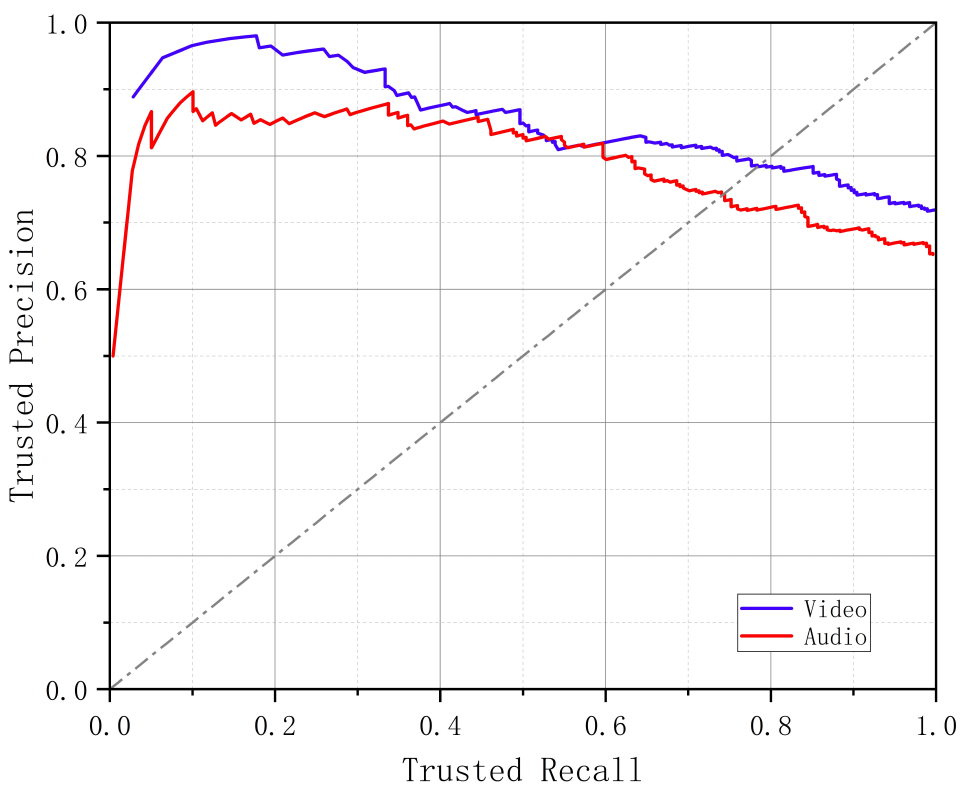}
    \caption{\textbf{The Trusted P-R curves of video and audio branchs, which is constructed by the sets of ($\mathrm{TR}$, $\mathrm{TP}$) based on confidence.}}
    \label{p_r}
\end{figure}

According to the analysis of Eq.\eqref{eq_tp} and Eq.\eqref{eq_tr}, we can deduce the relationship between the confidence threshold and trusted precision and trusted recall. When the confidence threshold increases, $\mathrm{HT}$, $\mathrm{HF}$, and $\mathrm{HT+HF}$ will decrease, $\mathrm{LT}$ will increase, and $\mathrm{HT+LT}$ will remain unchanged. In this case, trusted precision may fluctuate slightly, but trusted recall will significantly decrease. Conversely, when the confidence threshold decreases, $\mathrm{HT}$, $\mathrm{HF}$, and $\mathrm{HT+HF}$ will increase, $\mathrm{LT}$ will decrease, and $\mathrm{HT+LT}$ will remain unchanged. In this case, trusted precision may fluctuate slightly, but trusted recall will significantly increase. Excellent overall model performance requires balancing trusted precision and trusted recall, necessitating selecting a suitable confidence threshold. A set of data on the $\mathrm{(TR, TP)}$ based on confidence can be obtained by adjusting different confidence thresholds. This data can be used to construct a Trusted P-R curve, as shown in Fig.\ref{p_r}. To balance trusted precision and trusted recall, the $\mathrm{y = x}$ line was introduced, and the threshold corresponding to the intersection point with the P-R curve can be selected as the confidence threshold for the training model. The confidence threshold is an essential indicator for evaluating the performance of a trusted model. The lower the confidence threshold of a model, the higher its overall confidence.

\subsubsection{Trusted F1 Score}

In the common classification models, we can use the harmonic mean of precision and recall to measure the model performance. In the trusted model, trusted precision and trusted recall are equally essential performance metrics. Trusted precision and trusted recall are considered equally important in the trusted model, so we use a coefficient of 1 to average them harmonically. The definition of trusted F1 score is as follows:
\begin{equation}
    \mathrm{Trusted} \quad \mathrm{F1} = 2 \times \frac{\mathrm{TP} \times \mathrm{TR}}{\mathrm{TP} + \mathrm{TR}},
\end{equation}
the Trusted F1 score requires the model to achieve a balance between Trusted Precision and Trusted Recall, making it particularly useful for scenarios with imbalanced test samples. It not only improves the recognition rate of samples under high-confidence conditions but also increases the proportion of high-confidence correct predictions among all correctly identified samples. A higher F1 score indicates better overall performance of the trusted model.

\subsubsection{Trusted Accuarcy}

For classification models, the most basic performance evaluation metric is classification accuracy, which measures the ratio of correctly classified instances to the total number of test examples. However, for trusted models, we need to consider the impact of confidence on the classification results. Therefore, the concept of trusted accuracy is proposed, which measures the ratio of correctly classified instances with high confidence to the total number of instances classified with high confidence. The trusted accuracy is defined as follows:
\begin{equation}
    \mathrm{Trusted} \quad \mathrm{Accuracy} = \frac{\mathrm{HT}}{\mathrm{HT} + \mathrm{HF}},
\end{equation}
the trusted accuracy allows us to consider both classification accuracy and confidence when evaluating the performance of the trusted model. A higher trusted accuracy indicates that the classification results of the model have higher confidence.

\section{Experiments}
\label{experiments}

To evaluate the performance of the proposed TER model, we first evaluate the overall performance of the model using the trusted evaluation criterion proposed in Section \ref{t_evaluation}. Additionally, we introduce the accuracy (Acc.) and F1 score to evaluate the classification performance of the TER model. In this section, we first introduce the datasets utilized for model training and testing, as well as the preprocessing methods and the training details of the model. Subsequently, we conduct ablation studies on TER model to validate the effectiveness of the multimodal methods, including the single-modal, fusion, and loss methods. Finally, we compare the performance of the proposed TER model and the confidence of the predictions with the state-of-the-art methods on two datasets. 

\subsection{Datasets and Implementation Details}

IEMOCAP\textsuperscript{\cite{busso2008iemocap}} is the earliest established multimodal emotion recognition dataset, which includes 10 individuals (5 males and 5 females) performing spontaneous dialogues based on pre-designed scripts. These performances cover emotions from 10 topics such as neutral, happiness, sadness, etc. The dataset records multiple modalities of information including video, audio, facial motion capture, and textual transcriptions, with an average duration of 4.5 seconds for emotion segments. In our experiment, we selected data from six balanced classes, namely neutral, happy, sad, anger, frustration, and excited. Music-video\textsuperscript{\cite{pandeya2021deep}} is a dataset composed of music videos, containing 3,438 emotional segments from different music videos. The average duration of each segment is 30 seconds. The dataset has coarse-grained discrete labels: exciting, fear, neutral, relaxation, sad, and tension. The recorded data includes both video and audio, and we utilized all available data formats provided by the dataset for model training.

We performed preprocessing for each data type in the datasets to get the consistent input format. For all datasets, we extracted a frame sequence containing 32 frames from the videos as the input for the video module. During preprocessing, we randomly sampled frames from the videos to ensure data diversity and differences between adjacent frames. We randomly selected a starting frame and added frames to the sequence at a certain interval (in our experiment, we selected one frame every four frames as input data). To meet the model input requirements, for video data, we cropped the frames and applied data augmentation strategies such as flipping and normalization. For audio data, both datasets provide the audio data as the input for the audio module. We extracted Mel spectrogram features from the audio files, with a dimension of $3 \times 96 \times 64$ per second. The duration of the input data for the audio module is consistent with that of the video module. In our experiment, we accept 4 seconds of Mel spectrogram features as input features and padded them with 0 to address cases where the audio is shorter.

The proposed TER model is implemented in PyTorch\textsuperscript{\cite{paszke2019pytorch}}. We deploy the pre-trained parameters of Video Swin-Transformer and VGGish to initialize the model, and deploy the AdamW\textsuperscript{\cite{kingma2014adam}} optimizer with a cosine decay learning rate scheduler and 2.5 epochs of linear warmup. The inital learning rate is 1$e$-4. All models are trained on a Nvidia Tesla V100 with 32GB. Due to the memory constraint, we set the batch size to 8. To allow the model to learn global features and reduce the risk of overfitting, we use the delayed update strategy, where a backward pass and gradient update are performed every 4 batches. We would stop the training and report the best result when the training reached 30 epochs or the performance does not improve for more than 10 epochs.

\subsection{Ablation Studies}

\subsubsection{Validity of Submodules}

To evaluate the effectiveness of each submodule in our proposed TER model, we designed two variant models:

\begin{itemize}
    \item Only Video: This model only utilizes the video and confidence modules without incorporating the audio module.

    \item Only Audio: This model only utilizes the audio and confidence modules without incorporating the video module.
\end{itemize}
These variant models are trained in the onece training process as the TER model and generate outputs from both variant models and the TER model during testing. With this design, we can ensure the evaluation of each submodule's effectiveness within the same model and highlight the effectiveness of the multimodal.

The experimental results of the dissolution research are presented in Table \ref{Submodules}. The results indicate that the proposed TER model outperforms all the variant models in performance. This suggests that incorporating multimodal information in AVCA can enhance the overall model performance. We observe that on the IEMOCAP dataset, our TER model performs mostly better than the only video model but slightly lower in trusted accuracy. It could be attributed to the average duration of emotion segments provided by the dataset being 4 seconds, where shorter audio information cannot meet the training requirements of the model. During multimodal fusion, sufficiently trained video predictions dominate the final outcome, while insufficiently trained audio predictions negatively impact the performance. Our combining beliefs module utilizes the confidence value from each modality's results to suppress this negative impact and mitigate performance degradation. Incorrect results are assigned lower confidence values in the confidence module, whereas correct results are given higher confidence values. When combining the two results according to the DS-Combination Rule, the correct results carry a greater weight to ensure the final result's confidence. Measuring the overall trusted performance of the model should consider the Trusted F1 score. On the IEMOCAP dataset, the TER model's Trusted F1 score is significantly higher than all of the variant models, validating our analysis. The performance on the Music-video dataset also validates the effectiveness of the multimodal approach. The sub-modules that learn comprehensive features can mutually enhance each other during fusion, resulting in superior multimodal performance. The TER model outperforms both variant models across all performance metrics. With an average duration of emotion segments of 30 seconds in the Music-video dataset, the data augmentation scheme employed in our preprocessing becomes more effective. The model can learn more diverse feature information from different segments, which is the main reason for the superior performance of the TER model on this dataset.

\subsubsection{Validity of Fusion Module}

To evaluate the effectiveness of the combining beliefs module, we designed two commonly used early fusion and late fusion methods as variant methods:
\begin{itemize}
    \item Early Fusion: This method concatenates the features extracted by the two submodules and then performs dimensionality reduction and classification using a classifier.

    \item Late Fusion: This method directly combines the output results of the classifiers from the two submodules to obtain the final result.
\end{itemize}
Both variant methods utilize the same parameter initialization as the TER model. They are trained on the Music-video using a cosine decay learning rate scheduler and 2.5 epochs of linear warmup. During the testing phase, the output results of the variant methods are fed into the confidence module to obtain trusted results for evaluating the trusted performance of the variant methods.
\begin{table*}[t]
    
    \caption{\textbf{The ablation experiments of our proposed TER model in terms of Acc., Macro F1, Weighted F1, Trusted Acc., and Trusted F1 on the two commonly used dataset IEMOCAP and Music-video. Bold numbers indicate the best performance.}}
    \label{Submodules}
    
    \begin{center}
        \begin{tabular}{ccccccc}
            
            \toprule
            
            Dataset & Method & Acc. & Macro F1 & Weighted F1 & Trusted Acc. & Trusted F1\\
            
            \midrule
            
            ~ & Only Video & 0.5619 & 0.5416 & 0.5614 & 0.6314 & 0.6206\\
            
            IEMOCAP & Only Audio & 0.4820 & 0.4448 & 0.4766 & 0.5923 & 0.5978\\ 
            
            ~ & \textit{TER(ours)} & \textbf{0.6014} & \textbf{0.5725} & \textbf{0.6007} & \textbf{0.6514}  & \textbf{0.7511}\\
            
            \midrule
            
            ~ & Only Video & 0.7143 & 0.7111 & 0.7120 & 0.7786 & 0.7786\\
            
            Music-video & Only Audio & 0.6429 & 0.6377 & 0.6413 & 0.7205 & 0.7233\\ 
            
            ~ & \textit{TER(ours)} & \textbf{0.8130} & \textbf{0.8231} & \textbf{0.8232} & \textbf{0.8240} & \textbf{0.9035}\\
            
            \bottomrule
            
        \end{tabular}
    \end{center}
\end{table*}
\begin{table*}[t]
    
    \caption{\textbf{The performance comparison of early fusion and late fusion in terms of Acc., Macro F1, Weighted F1, Trusted Acc., and Trusted F1 on the Music-video. Bold numbers indicate the best performance.}}
    \label{Fusion}
    
    \begin{center}
        \begin{tabular}{cccccc}
            
            \toprule
            
            Fusion Method & Acc. & Macro F1 & Weighted F1 & Trusted Acc. & Trusted F1\\
            
            \midrule
            
            Early Fusion & 0.7806 & 0.7795 & 0.7821 & 0.8001 & 0.8231\\
            
            Late Fusion & 0.7806 & 0.7800 & 0.7806 & 0.7941 & 0.8155\\ 
            
            \textit{Combining beliefs(ours)}  & \textbf{0.8130} & \textbf{0.8231} & \textbf{0.8232} & \textbf{0.8240} & \textbf{0.9035}\\
            
            \bottomrule
            
        \end{tabular}
    \end{center}
\end{table*}

The performance comparison of different fusion methods is shown in Table \ref{Fusion}. The experimental results demonstrate that our combining beliefs module outperforms both early fusion and late fusion. Compared to early fusion, our fusion method is more interpretable. Early fusion concatenates features and performs convolution, inheriting the lack of interpretability of deep models. On the other hand, our fusion method combines different classification results based on their confidence value, using the Dempster-Shafer theory to ensure the rationality and effectiveness of fusion. Confidence values are computed using the confidence module based on variational Dirichlet. Our fusion method shares similarities with late fusion in using weighting coefficients to combine multiple classification results but with a distinctive feature. Traditional late fusion methods often seek the weighting coefficients that optimize the overall performance. In contrast, our fusion method finds the most suitable fusion weights for each group of classification results based on their confidence values, resulting in dynamic fusion weights. Therefore, our fusion method addresses the limitations of existing fusion methods, offering stronger theoretical foundations and interpretability. The dynamic fusion weights enable us to find the optimal fusion strategy for each fusion, thus yielding superior classification performance and confidence compared to existing fusion methods.

\subsubsection{Validity of Trusted CE Loss}

To evaluate the effectiveness of the proposed Loss methods, we have designed commonly used loss calculation methods, as well as several variations incorporating confidence value:
\begin{itemize}
    \item \textbf{CE:} Calculating the cross-entropy loss. \\ $\mathcal{L}(\hat{Y}_i; Y_i) = -\sum_{c=1}^C y_{ic} \log \hat{y}_{ic}$

    \item \textbf{Add Trusted:} Incorporating credibility into the cross-entropy calculation. \\ $\mathcal{L}(\hat{Y}_i, \hat{U}_i; Y_i) = -\sum_{c=1}^C y_{ic} \log \hat{y}_{ic} + \hat{u_i}$

    \item \textbf{Tan(Mul) Trusted:} Combining confidence value with the tan function curve to enhance its impact on the results. \\ $\mathcal{L}(\hat{Y}_i, \hat{U}_i; Y_i) = -\sum_{c=1}^C y_{ic} \log \hat{y}_{ic} \times \tan (\hat{u_i} \times \frac{\pi}{2})$

    \item \textbf{Tan(Add) Trusted:} Combining confidence value with the tan function curve, controlling overfitting through addition. \\ $\mathcal{L}(\hat{Y}_i, \hat{U}_i; Y_i) = -\sum_{c=1}^C y_{ic} \log \hat{y}_{ic} + \tan (\hat{u_i} \times \frac{\pi}{2})$

    \item \textbf{Exp(Mul) Trusted:} Combining confidence value with the exponential function curve to enhance its impact on the results. \\ $\mathcal{L}(\hat{Y}_i, \hat{U}_i; Y_i) = -\sum_{c=1}^C y_{ic} \log \hat{y}_{ic} + \exp (\hat{u_i})$
\end{itemize}
These loss calculation methods, including the proposed Trusted CE, are trained and tested using an end-to-end TER model with the sum of losses from audio, video, and fusion results. We have performed training and testing on the Music-video dataset with sufficient data volume to evaluate the performance of these loss calculation methods.

\begin{table*}[t]
    
    \caption{\textbf{The performance comparison of our proposed Trusted CE and other loss functions in terms of Acc., Macro F1, Weighted F1, Trusted Acc., and Trusted F1 on the Music-video. Bold numbers indicate the best performance.}}
    \label{Loss}
    
    \begin{center}
    \begin{tabular}{cccccc}
        \toprule
        Loss Method & Acc. & Macro F1 & Weighted F1 & Trusted Acc. & Trusted F1\\
        \midrule
        CE & 0.7581 & 0.7729 & 0.7742 & 0.7781 & 0.8230\\
        Add Trusted & 0.5432 & 0.4935 & 0.5156 & 0.5536 & 0.7126\\ 
        Tan(Mul) Trusted & 0.1645 & 0.0494 & 0.0514 & 0.1735 & 0.2957\\ 
        Tan(Add) Trusted & 0.5232 & 0.4807 & 0.4979 & 0.5332 & 0.6955\\ 
        Exp Trusted & 0.5110 & 0.4801 & 0.4882 & 0.5128 & 0.6779\\ 
        \textit{Trusted CE(ours)}  & \textbf{0.8130} & \textbf{0.8231} & \textbf{0.8232} & \textbf{0.8240} & \textbf{0.9035}\\
        \bottomrule
    \end{tabular}
\end{center}
\end{table*}
\vspace{3mm} 

\begin{figure}[h]
    \centering
    \includegraphics[width=\linewidth]{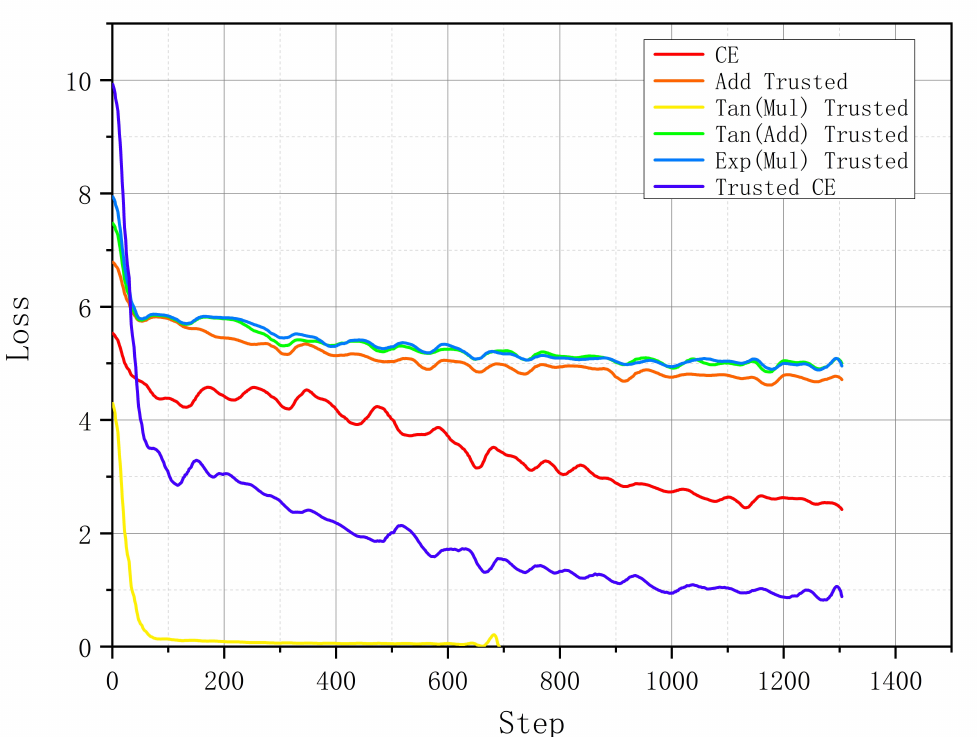}
    \caption{\textbf{The loss comparison of our proposed Trusted CE and another four loss functions during the training process. The Tan(Mul) Trusted method
exhibits a rapid decrease in loss. It was terminated at 700 steps due to its poor performance. Trusted CE ranks second.}}
    \label{loss}
\end{figure}

\begin{figure}[h]
    \centering
    \includegraphics[width=1\linewidth]{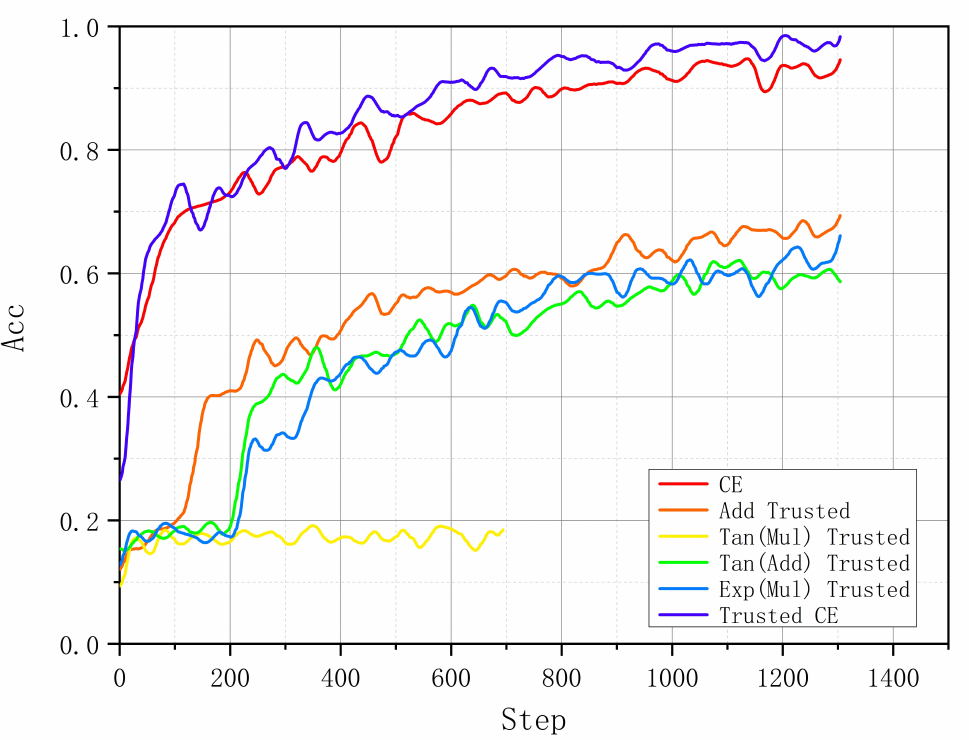}
    \caption{\textbf{The performance comparison of our proposed Trusted CE and another four loss functions in terms of Acc. during the training process. The performance of the Tan(Mul) Trusted method exhibits only minimal fluctuations. It was terminated at 700 steps due to its poor performance. Trust CE achieves the highest accuracy as the number of steps increases. }}
    \label{acc}
\end{figure}

The changes in loss during the training process are depicted in Fig. \ref{loss}, while the changes in accuracy are shown in Fig. \ref{acc}. From Fig. \ref{loss} and Fig. \ref{acc}, it is evident that the Tan(Mul) Trusted method shows a rapid decrease in loss, but its accuracy does not significantly improve with an increase in training steps. However, Trusted CE ranks second in loss reduction, yet it exhibits a notable improvement in accuracy compared to other loss functions. Therefore, Trusted CE is the best method for loss calculation.

The performance comparison of different methods is presented in Table \ref{Loss}. From the figures, we first notice that the Tan(Mul) Trusted method exhibits a rapid decrease in loss, followed by a stable trend, while the model's performance shows only minimal fluctuations without significant improvement. This outcome primarily attributes to the choice of the loss function. The curve of the tan function allows the model to fit the minimum loss value quickly, but its impact on model performance is limited, resulting in the obtained model falling short of being effective in classification. The Add Trusted, Tan(Add) Trusted, and Exp(Mul) Trusted methods exhibit similar trends in the training loss, with similar accuracy and final performance. We observe that simply adding confidence value to the loss function affects the speed of loss decrease, and the range of loss variation in the classification part also decreases, leading to decreased information learned by the model and an inability to achieve excellent performance after stabilization. After observing the training curves of different calculation methods, we incorporate confidence value into the loss function using the cross-entropy-based approach. From the training process, we can observe that our proposed Trusted CE exhibits faster loss reduction, lower final fitted loss, quicker improvement in model performance, and results in the TER model's best trusted performance.

\subsection{Comparison with SOTA Methods}

This sub-section performs the AVCA on the IEMOCAP and Music-video datasets. The results on the validation sets for the proposed TER model compared with the state-of-the-art methods are shown as Table \ref{IEMOCAP} and Table \ref{Music-video}, respectively. Bold font indicates the best score for experiments with different evaluation criteria.

\begin{table*}[t]
    
    \caption{\textbf{The performance comparison of our proposed TER model and other state-of-the-art methods in terms of Acc., Macro F1, Weighted F1, Trusted Acc., and Trusted F1 on the IEMOCAP. Bold numbers indicate the best performance.}}
    \label{IEMOCAP}
    
    \begin{center}
    \begin{tabular}{cccccc}
        \toprule
        Method & Acc. & Macro F1 & Weighted F1 & Trusted Acc. & Trusted F1\\
        \midrule
        bc-LSTM\textsuperscript{\cite{poria2017context}} & 0.5860 & 0.5725 & 0.5860 & 0.5612 & 0.5642\\
        DialogueGCN\textsuperscript{\cite{ghosal2019dialoguegcn}} & 0.6057 & 0.5719 & 0.5868 & 0.5641 & 0.5742\\ 
        DAG-ERC\textsuperscript{\cite{shen2021directed}} & \textbf{0.6794} & \textbf{0.6682} & \textbf{0.6782} & 0.5838 & 0.5921\\ 
        CIM\textsuperscript{\cite{akhtar2019multi}} & 0.5693 & 0.5478 & 0.5614 & 0.5230 & 0.5233\\ 
        COGMEN\textsuperscript{\cite{joshi2022cogmen}} & 0.6118 & 0.6097 & 0.6136 & 0.5521 & 0.5520\\ 
        MMGCN\textsuperscript{\cite{hu2021mmgcn}} & 0.6562 & 0.6411 & 0.6536 & 0.5424 & 0.5415\\ 
        \textit{TER(ours)} & 0.6014 & 0.5725 & 0.6007 & \textbf{0.6514} & \textbf{0.7511}\\
        \bottomrule
    \end{tabular}
\end{center}

\end{table*}

\begin{figure}[h]
    \centering
    \includegraphics[width=\linewidth]{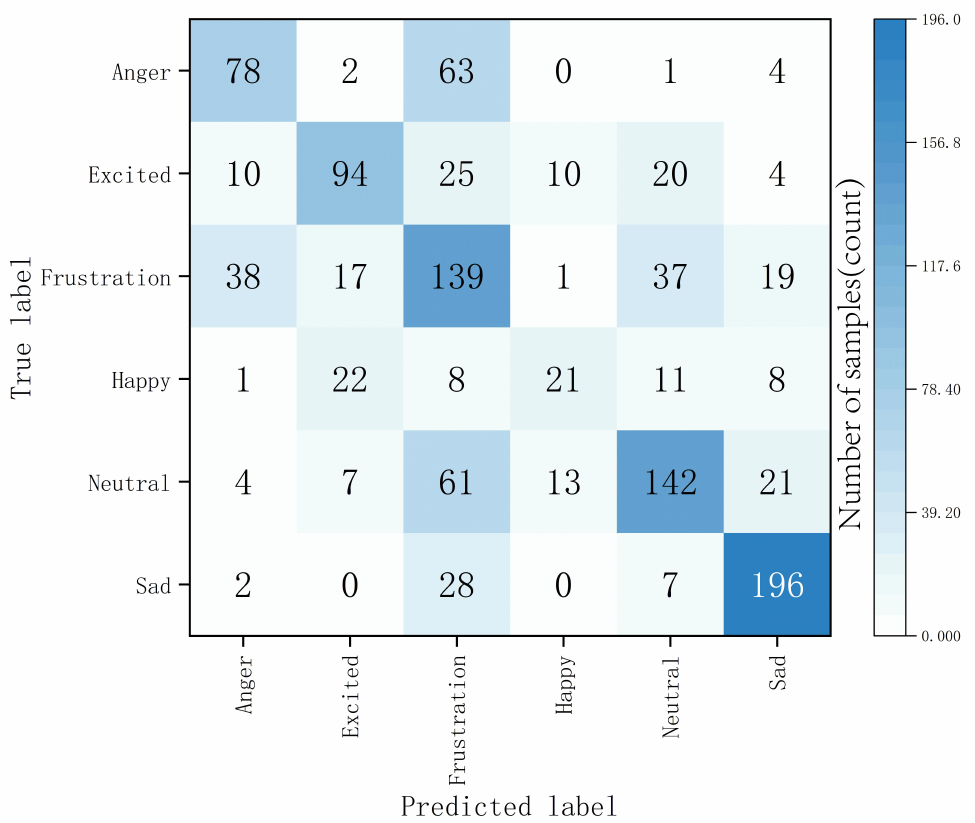}
    \caption{\textbf{The confusion matrix of our proposed TER model on the IEMOCAP dataset.}}
    \label{iemocap}
\end{figure}

\begin{figure}[h]
    \centering
    \includegraphics[width=\linewidth]{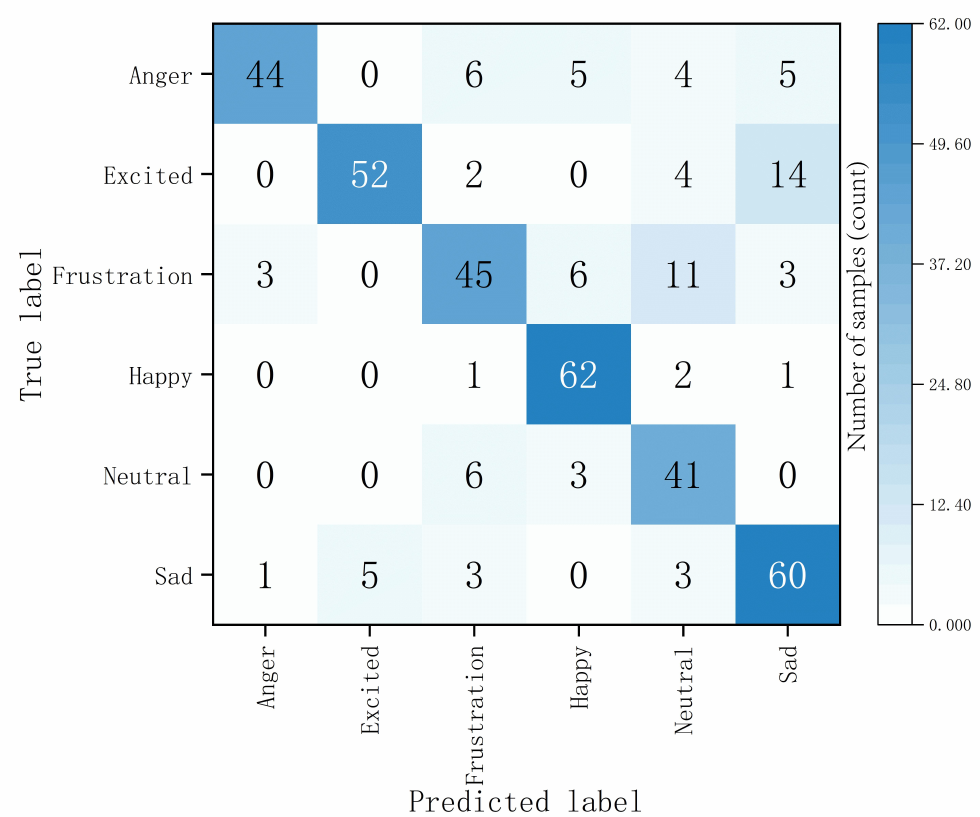}
    \caption{\textbf{The confusion matrix of our proposed TER model on the Music-video dataset.}}
    \label{music_video}
\end{figure}

We first present the best classification confusion matrix of the TER model on the IEMOCAP dataset, as shown in Fig.\ref{iemocap}. Our model achieves excellent classification performance when the training data is sufficient. However, when the training data is insufficient, the model fails to learn effective features. This observation is validated by the training results for the ``happy" label, which exhibits significantly lower classification performance due to the limited amount of available data. Compared to the state-of-the-art model performance shown in Table \ref{IEMOCAP}, these high-performing models utilize manually designed feature information as model inputs, whereas our TER model extracts features by the deep model. Under this premise, our proposed TER model achieves classification performance close to the SOTA methods, demonstrating its stronger feature extraction capability. On the other hand, we emphasize the model's trusted performance, which is an important innovation of our method. We make minor modifications to the SOTA models to evaluate their trusted performance. During testing, we change their softmax layer to the softplus layer and input its result into our confidence module to obtain trusted results, then evaluated for trusted performance. We can observe that our TER model achieves SOTA trusted performance in both trusted accuracy and trusted F1 score. We notice that the trust accuracy of all SOTA models is lower than their accuracy, and the same applies to their F1 scores. However, our TER model not only achieves trusted performance equal to or even higher than its classification performance but also exhibits a significantly higher trusted F1 score than the F1 score. This can be attributed to our proposed Trusted CE Loss, which incorporates confidence value into the loss computation to ensure the model's trusted performance and prevent the model from overfitting to classification performance and neglecting trusted performance during model training. Our combining beliefs module also guarantees the confidence value of the fused results. These proposed trusted methods synergistically enhance the model's trusted performance, enabling TER to achieve SOTA trusted performance on the IEMOCAP dataset.
\begin{table*}[t]
    
    \caption{\textbf{The performance comparison of our proposed TER model and other state-of-the-art methods in terms of Acc., Macro F1, Weighted F1, Trusted Acc., and Trusted F1 on the Music-video. Bold numbers indicate the best performance.}}
    \label{Music-video}
    
    \begin{center}
    \begin{tabular}{cccccc}
        \toprule
        Method & Acc. & Macro F1 & Weighted F1 & Trusted Acc. & Trusted F1\\
        \midrule
        CIM\textsuperscript{\cite{akhtar2019multi}} & 0.4913 & 0.4787 & 0.4856 & 0.4828 & 0.4836\\ 
        COGMEN\textsuperscript{\cite{joshi2022cogmen}} & 0.5223 & 0.5079 & 0.5232 & 0.5041 & 0.5040\\ 
        MMGCN\textsuperscript{\cite{hu2021mmgcn}} & 0.5526 & 0.5497 & 0.5502 & 0.5156 & 0.5150\\ 
        \textit{TER(ours)} & \textbf{0.8130} & \textbf{0.8231} & \textbf{0.8232} & \textbf{0.8240} & \textbf{0.9035}\\
        \bottomrule
    \end{tabular}
\end{center}

\end{table*}

Next, we demonstrate the best performance the TER model can achieve with sufficient data. The Fig. \ref{music_video} shows the best classification confusion matrix of the TER model on the Music-video dataset. Although the Music-video dataset has fewer videos than IEMOCAP, each video has a longer duration, resulting in more diverse training samples after preprocessing. The TER model exhibits its powerful feature extraction capability and classification performance on the Music-video dataset, showing similar high classification performance across different labels. We compare the performance of the SOTA methods with the TER model by transferring them to the Music-video dataset. Since the Music-video dataset does not provide text information, we compare three multimodal models, disable the model's text feature input, and modify specific parameters to complete the model training. We extract the emotional features of the dataset as model inputs according to the requirements of each model using OpenSmile\textsuperscript{\cite{eyben2010opensmile}}. The detailed performance comparison of the proposed TER model and SOTA methods on the Music-video dataset are shown in Table \ref{Music-video}. First of all, we notice that the performance of the three SOTA methods all decreases due to the disabling of the text feature input. The loss of many effective features inevitably leads to a degradation in model performance. In contrast, our TER model remains unaffected and exhibits even better classification and trusted performance due to increased data volume. Therefore, a large-scale dataset is indispensable if one wishes to train an end-to-end deep model with high performance. Finally, through research on multimodal information and fusion rules, we find that multimodal information can enhance the trusted performance of the overall model. High-confidence modalities can compensate for the shortcomings of low-confidence modalities and improve the overall trusted performance of the model. This advantage of multimodal information in confidence is superior to single-modal information.

\subsection{Improve Trusted Performence}

This sub-section discusses the generality of the proposed Trusted CE Loss method. We have conducted comparative experiments on the IEMOCAP dataset to validate the effectiveness and generality of this method in improving trusted performance. 

\begin{table*}[t]
    
    \caption{\textbf{The performance comparison of the state-of-the-art methods and their trusted ones in terms of Acc., Macro F1, Weighted F1, Trusted Acc., and Trusted F1 on the IEMOCAP. Bold numbers indicate better performance of the six existing methods and their trusted variants across six evaluation metrics.}}
    \label{Improve}
    
    \begin{center}
        \begin{tabular}{ccccccc}
    \toprule
    Method & Acc. & Macro F1 & Weighted F1 & Trusted Acc. & Trusted F1 & Trusted Threshold\\
    \midrule
    bc-LSTM\textsuperscript{\cite{poria2017context}} & \textbf{0.5860} & \textbf{0.5725} & \textbf{0.5860} & 0.5612 & 0.5642 & 0.5341\\
    Trusted bc-LSTM & 0.5798 & 0.5720 & 0.5826 & \textbf{0.8114} & \textbf{0.8210} & \textbf{0.4608}\\
    DialogueGCN\textsuperscript{\cite{ghosal2019dialoguegcn}} & 0.6057 & 0.5719 & 0.5868 & 0.5641 & 0.5742 & 0.6481\\ 
    Trusted DialogueGCN & \textbf{0.6081} & \textbf{0.5759} & \textbf{0.5870} & \textbf{0.6180} & \textbf{0.6280} & \textbf{0.4620}\\ 
    DAG-ERC\textsuperscript{\cite{shen2021directed}} & \textbf{0.6794} & \textbf{0.6682} & \textbf{0.6782} & 0.5838 & 0.5921 & 0.6520\\ 
    Trusted DAG-ERC & 0.6615 & 0.6528 & 0.6619 & \textbf{0.7236} & \textbf{0.7245} & \textbf{0.4647}\\ 
    CIM\textsuperscript{\cite{akhtar2019multi}} & \textbf{0.5693} & \textbf{0.5478} & \textbf{0.5614} & 0.5230 & 0.5233 & 0.5694\\ 
    Trusted CIM & 0.5619 & 0.5379 & 0.5414 & \textbf{0.6101} & \textbf{0.6155} & \textbf{0.4681}\\ 
    COGMEN\textsuperscript{\cite{joshi2022cogmen}} & 0.6118 & 0.6097 & 0.6136 & 0.5521 & 0.5520 & 0.5360\\ 
    Trusted COGMEN & \textbf{0.6420} & \textbf{0.6350} & \textbf{0.6438} & \textbf{0.7087} & \textbf{0.7048} & \textbf{0.4399}\\
    MMGCN\textsuperscript{\cite{hu2021mmgcn}} & \textbf{0.6562} & \textbf{0.6411} & \textbf{0.6536} & 0.5424 & 0.5415 & 0.7378\\ 
    Trusted MMGCN & 0.6451 & 0.6213 & 0.6399 & \textbf{0.6714} & \textbf{0.6736} & \textbf{0.4231}\\
    \bottomrule
\end{tabular}

    \end{center}
\end{table*}

\begin{table}[ht]
\centering
\captionsetup{skip=10pt}  
\caption{Computational Cost Comparison With Different Methods 
(In Terms of GFLOPs, Number of Parameters and Runtime).
}
\vspace{10pt}  
\label{table:cost_comparison}
\begin{tabular}{@{}lccc@{}}
\toprule
Method     & GFLOPs & Parameters & Runtime \\ 
\midrule
Late Fusion  & 74.36   & 316.07 $\times 10^6$ & 59.4 h \\
 \textit{Combining beliefs(ours)}  & 74.36  & 316.07 $\times 10^6$ & 63.6 h \\
\bottomrule
\end{tabular}
\end{table}

We apply the trusted improvement to each SOTA model to obtain the trusted-improved model. The trusted improvement includes model output transformation, calculating confidence value of predictions, and trusted loss calculation. First of all, we replace the softmax layer in the model output layer with a softplus layer and pass its output to the confidence module to obtain trusted results with a specific confidence value. Then, we calculate the trusted loss according to Eq.\eqref{eq_loss} and perform model training through backpropagation. The trusted-improved model's results are tested for classification and trusted performance during model testing.

The performance comparison between SOTA and trusted-improved models is presented in Table \ref{Improve}. Bold font indicates that the trusted improvement model achieved performance enhancement compared to the original model for that evaluation criterion. 
First, we observe that in terms of trusted evaluation metrics, all the models improved with trustworthiness outperform the original models, effectively validating the generality of our proposed method. We also compared the GFLOPs, parameters, and runtime of our method with those of the late fusion approach on the IEMOCAP dataset, as shown in Table \ref{table:cost_comparison}. In Table \ref{table:cost_comparison}, we found that while our method matches the late fusion approach in terms of GFLOPs and parameters, it requires slightly more training time (30 epochs). However, as shown in Table \ref{Improve}, the performance improvement of our method is significant, with an average increase in trusted performance of 20\%, further confirming the effectiveness of our approach. On the other hand, in terms of classification performance,
trusted-improved models exhibit similar performance to the original models, and even the trust-improved models of DialogueGCN\textsuperscript{\cite{ghosal2019dialoguegcn}} and COGMEN\textsuperscript{\cite{joshi2022cogmen}} outperform the original models in classification performance, overall surpassing them. 
Our method enhances the overall performance of the models, ensuring the preservation of the classification performance while significantly improving the trusted performance.

\section{Conclusion}
\label{conclusion}

Despite the superior performance in emotion classification demonstrated by previous studies\textsuperscript{\cite{shen2021directed, hu2021mmgcn}}, their limitations in practical applications are evident. These models rely on manually designed emotion features as input, different from mainstream end-to-end neural networks that can automatically extract features from raw data. These models suffer from more severe performance degradation issues when dealing with data that does not include textual information. To address the challenges posed by existing models, we propose an end-to-end framework called TER for AVCA. The TER model trains the emotion feature extractor and classifier in the end-to-end manner, eliminating the needs for manual feature extraction. When sufficient training data is available, TER exhibits superior classification performance compared to existing SOTA models. Even on the IEMOCAP dataset, TER achieves comparable performance to SOTA models. Therefore, the proposed TER model possesses excellent feature extraction and classification capabilities, and its end-to-end structure is more conducive to practical implementation and deployment.

The model's trusted performance is another significant contribution, and we realize a trusted deep learning framework by introducing the confidence method proposed by Han et al.\textsuperscript{\cite{han2022trusted}}. Previous research has explored confidence methods but needed more effective evaluation criteria, resulting in limited research and application of confidence methods. This paper proposes a series of trusted evaluation criteria to address this issue and evaluates existing SOTA models and proposed TER model to showcase the advancement of trusted models. In existing methods, model training focuses solely on classification performance, leading to lower trusted performance. However, in the training process of TER, we incorporate confidence value into the loss calculation method, thereby balancing both aspects of performance. The Trusted CE Loss plays a crucial role in improving the confidence of existing methods, significantly enhancing trusted performance while maintaining classification performance. Trusted models, trusted CE loss, and trusted evaluation criterion constitute our trusted approach, and extensive ablation and comparative experiments validate its effectiveness and superiority.

In future work, larger datasets can be utilized to enhance model performance and generalization, such as the MELD dataset\textsuperscript{\cite{poria2018meld}}, and better performance can be achieved on smaller datasets through transfer learning.
Second, more modalities will be added because the flexible combining beliefs module will provide more evidence to improve classification and trusted performance. Finally, using the knowledge distillation method to reduce the parametric number of the model and investigate models with smaller parametric numbers is also an efficient solution.

\section{Acknowledgment}
\label{acknowledgement}

\noindent
The work of this paper is supported by the Key R\&D Program of Zhejiang Province (No. 2023C01181). 



\vskip 2mm

\renewcommand\refname{\large\textbf{References}}

\bibliographystyle{IEEEtran}
\bibliography{IEEEabrv,mylib}

\mbox{}
\clearpage
\clearpage
\large
\end{document}